\documentclass[10pt,twocolumn,letterpaper]{article}

\usepackage[pagenumbers]{cvpr} 

\definecolor{cvprblue}{rgb}{0.21,0.49,0.74}

\definecolor{tabfirst}{rgb}{1, 0.7, 0.7}
\definecolor{tabsecond}{rgb}{1, 0.85, 0.7}
\definecolor{tabthird}{rgb}{1, 1, 0.7}
\newcommand{\cellfirst}[1]{\cellcolor{tabfirst}{#1}}
\newcommand{\cellsecond}[1]{\cellcolor{tabsecond}{#1}}
\newcommand{\cellthird}[1]{\cellcolor{tabthird}{#1}}

\usepackage[pagebackref,breaklinks,colorlinks,allcolors=cvprblue]{hyperref}

\usepackage[english]{babel}
\usepackage{colortbl}
\usepackage{xcolor} 
\usepackage{tabularx}
\usepackage{multirow}
\usepackage{booktabs}
\usepackage{graphicx}
\usepackage{array}
\usepackage{caption}
\usepackage{amsmath}
\usepackage{float}
\usepackage{booktabs}
\usepackage{capt-of}
\usepackage[accsupp]{axessibility}  

\def\paperID{40058} 
\def\confName{CVPR}
\def\confYear{2026}

\title{RT-Splatting: Joint Reflection-Transmission Modeling with Gaussian Splatting}

\author{
    Ji Shi\quad
    Xianghua Ying$^*$\quad
    Bowei Xing\quad
    Ruohao Guo\quad
    Wenzhen Yue
    \vspace{0.2cm}
\\
    State Key Laboratory of General Artificial Intelligence\\
    School of Intelligence Science and Technology\\
    Peking University
}

\begin{document}
\twocolumn[{
\maketitle
\begin{center}
\captionsetup{type=figure}
\includegraphics[width=1.0\textwidth]{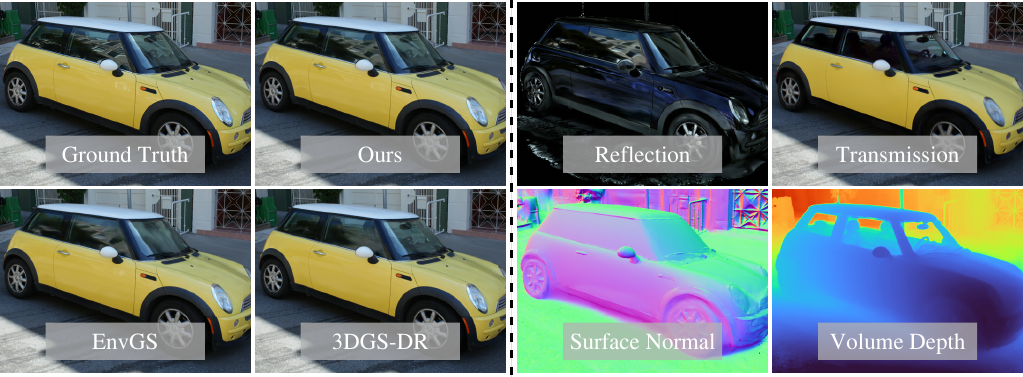}
\caption{\textbf{Photorealistic rendering and decomposition of real-world scene with coexisting reflection and transmission.} (Left) Compared to prior works, our method robustly handles semi-transparent surfaces, avoiding blurry reflections or overly occluded transmission. (Right) Our high-fidelity results are achieved by decomposing the scene radiance into Reflection and Transmission layers, enabled by a unified Gaussian representation that jointly captures surface geometry and scene volume.}
\label{fig:teaser}
\end{center}
\bigbreak
}]

\renewcommand{\thefootnote}{\fnsymbol{footnote}}
\footnotetext[1]{Corresponding author.}

\begin{abstract}    
3D Gaussian Splatting (3DGS) enables real-time novel view synthesis with high visual quality. However, existing methods struggle with semi-transparent specular surfaces that exhibit both complex reflections and clear transmission, often producing blurry reflections or overly occluded transmission. To address this, we present \textbf{RT-Splatting}, a framework that disentangles each Gaussian's geometric occupancy from its optical opacity. This factorization yields a unified surface-volume scene representation with a single set of Gaussian primitives. Our hybrid renderer interprets this representation both as a surface to capture high-frequency reflections and as a volume to preserve clear transmission. To mitigate the ambiguity in jointly optimizing reflection and transmission, we introduce Specular-Aware Gradient Gating, which suppresses misleading gradients from highly specular regions into the transmission branch, effectively reducing distracting floaters. Experiments on challenging semi-transparent scenes show that RT-Splatting achieves state-of-the-art performance, delivering high-fidelity reflections and clear transmission with real-time rendering. Moreover, our factorization naturally enables flexible scene editing. The project page is available at \href{https://sjj118.github.io/RT-Splatting}{https://sjj118.github.io/RT-Splatting}.
\end{abstract}
\section{Introduction}
\label{sec:intro}

3D Gaussian Splatting (3DGS)~\cite{3dgs} has revolutionized the field of novel view synthesis with its real-time rendering capabilities, achieved by representing a scene as a sparse set of 3D Gaussian primitives and rendering them efficiently via rasterization. Despite its success, 3DGS struggles to model semi-transparent specular surfaces where reflection and transmission coexist. To reproduce high-frequency specular highlights, standard 3DGS often hallucinates ``floaters'' behind the surface. These behind-surface floaters not only fail to faithfully capture the true reflected appearance, but also corrupt transmission by spuriously occluding background geometry that should remain visible through the surface.

Recent 3DGS variants address high-frequency view-dependent effects by replacing per-Gaussian SH with physically based shading that explicitly evaluates the rendering equation using scene geometry, material properties, and incident illumination~\cite{gao2024relightable,prtgs,shi2025gir,gs-ir,gaussianshader}. Early work performed shading at the primitive level (per-Gaussian), whereas more recent methods adopt deferred shading~\cite{3dgs-dr,dihlmann2024subsurface,gi-gs,gs-id,wu2024deferredgs,ye2024progressive,gs-ror2,rgs-dr,yao2024reflective,geosplatting,spectre-gs,refgs,envgs}, which first rasterizes scene properties into G-buffers and then performs per-pixel shading. However, because the G-buffer stores only the properties of the nearest surface at each pixel, transparency is fundamentally difficult to handle in deferred shading~\cite{aladaugli2015deferred}. For semi-transparent specular surfaces exhibiting both reflection and transmission, these methods either fail to aggregate the target surface's attributes needed for reflection modeling or simply treat the surface as opaque, completely occluding transmission. TransparentGS~\cite{transparentgs} instead adopts a multi-stage pipeline, where transparent objects are modeled with a separate set of Gaussian primitives on top of a background reconstructed in a separate stage using standard 3DGS while masking out transparent regions. Because the background is reconstructed without seeing through the transparent objects, the method struggles in scenes where the background is exclusively visible through the transparent surfaces, such as viewing a car's interior solely through its windows.

In this paper, we present \textit{RT-Splatting}, a hybrid surface-volume rendering framework for jointly modeling reflection and transmission in real-world scenes containing thin semi-transparent surfaces. For each Gaussian primitive, we decouple its role as a surface element from its role in attenuating light along the ray. Specifically, we factorize its contribution into a geometric occupancy term and an optical opacity term, thereby enabling a unified surface-volume scene representation with a single set of Gaussians. The geometric occupancy determines how strongly the Gaussian participates as a surface element along a ray, while the optical opacity controls how much light is absorbed or scattered once that surface is hit. This unified representation naturally supports a hybrid rendering pipeline: geometric occupancy is used to aggregate first-hit surface attributes into G-buffers for deferred reflection shading, whereas optical opacity drives a volumetric forward pass that accumulates transmitted background radiance.

However, even with this surface-volume formulation, jointly optimizing reflection and transmission remains ambiguous. High-frequency specular reflections are inherently difficult to fit, and the residual errors tend to produce misleading gradients that leak into the transmission branch. This causes the transmission component to compensate by creating erroneous floaters that corrupt background clarity. To mitigate this issue, we introduce a \textit{Specular-Aware Gradient Gating} mechanism that identifies pixels dominated by complex specular patterns and attenuates the corresponding gradients flowing to the transmission branch. This gating suppresses misleading supervision, substantially reduces distracting floaters, and improves the clarity of the transmitted background.

To summarize, our contributions are as follows:
\begin{itemize}
    \item We introduce a unified surface-volume Gaussian scene representation for jointly modeling sharp specular reflections and clear transmission in real-world scenes containing thin semi-transparent surfaces.
    \item We propose Specular-Aware Gradient Gating to suppress misleading gradients from complex specular regions, substantially reducing floaters in the transmission branch.
    \item Extensive experiments demonstrate that RT-Splatting significantly outperforms prior methods while maintaining real-time rendering and enabling flexible scene editing.
\end{itemize}

\section{Related Work}
\label{sec:related}

\subsection{Reflective Scene Reconstruction}

The reconstruction and rendering of reflective scenes is a long-standing challenge in novel view synthesis. Ref-NeRF~\cite{refnerf} conditions outgoing radiance on the reflection direction, rather than the viewing direction, improving the capture of high-frequency specular effects. Subsequent works advance this idea by either strengthening directional encodings to better capture light-surface interactions~\cite{nmf,nero,envidr,specnerf} or recovering more accurate surface geometry~\cite{unisdf,normal-nerf,refneus} to mitigate shape-radiance ambiguity~\cite{nerf++}. To address the challenge of rendering consistent reflections of nearby content, NeRF-Casting~\cite{nerfcasting} performs cone tracing along reflection paths and aggregates features before decoding, yielding high-fidelity inter-reflections. However, these approaches rely on dense implicit-field queries along rays during both training and inference, making real-time rendering impractical.

In contrast, recent works leveraging Gaussian Splatting have achieved real-time rendering capabilities for reflective scenes. GaussianShader~\cite{gaussianshader} estimates per-Gaussian normals from the shortest axis and shades with a learnable environment map for efficient specular shading. 3DGS-DR~\cite{3dgs-dr} adopts a deferred pipeline that first rasterizes scene attributes into G-buffers and then performs per-pixel shading. Ref-GS~\cite{refgs} extends 2DGS~\cite{2dgs} with a directional factorization for spatio-angular view-dependent effects. EnvGS~\cite{envgs} further employs a differentiable Gaussian ray tracer with environment Gaussians to capture near-field reflections in real time. While these methods excel at representing high-frequency specular effects, they still struggle with thin, semi-transparent surfaces whose appearance is a mixture of light transmitted through the surface and light reflected from the surface.

\subsection{Transparent Object Reconstruction}

While the native alpha-blending in volumetric methods like NeRF~\cite{nerf} and 3DGS~\cite{3dgs} can simulate translucency, it does so by conflating the geometric presence of a surface with its optical transmissivity, preventing it from establishing a distinct surface geometry required for physically-based shading. To circumvent this ambiguity, one line of research~\cite{dex-nerf,clear-splatting,tsgs} explores various first-surface extraction strategies to explicitly recover the surface of transparent object. Other works focus on the challenging task of reconstructing the complex view-dependent appearance on the transparent surface. To make this highly ill-posed problem tractable, a predominant strategy involves decoupling the object from its environment. This is typically achieved either by employing multi-stage pipelines to pre-reconstruct and freeze the opaque background~\cite{transparentgs,bemana2022eikonal,gao2023transparent}, or by simplifying the background to an infinitely distant environment map~\cite{nerrf,nemto}. Such approaches, however, are not applicable to general, complex scenes where transparent object and diffuse background are photometrically entangled (\eg, when the background is only visible through the transparent surface). Some approaches further impose strong constraints on the scene configuration, such as assuming simplified geometry like planar surfaces~\cite{ref2-nerf} or requiring controlled capture conditions like forward-facing camera arrangements~\cite{nerfrac}.

The restrictions in these methods often stem from the inherent ill-posedness of disentangling reflection and refraction, since both phenomena are highly view-dependent and lack the multi-view photometric consistency. To avoid this challenge, a practical approach is to focus on ubiquitous thin semi-transparent surfaces, such as glass panes or plastic films, where the negligible refractive effect allows light transport to be approximated as straight-path transmission. Following this direction, recent works~\cite{gao2024planar,zhang2024refgaussian} have shown promise in jointly modeling reflection and transmission, but their applicability remains limited to simple planar surfaces, failing to generalize to complex shapes.

\section{Preliminaries}
\label{sec:preliminary}

\subsection{Gaussian Splatting}

3D Gaussian Splatting (3DGS) \cite{3dgs} has recently emerged as a powerful technique for real-time, high-fidelity novel view synthesis. It represents a 3D scene with a collection of anisotropic 3D Gaussian primitives, each defined by its position, covariance, opacity $\alpha$, and color represented by Spherical Harmonics (SH). During rendering, these 3D Gaussians are projected onto the 2D image plane and sorted by depth. The final color $\mathbf{C}$ for a pixel is then computed by alpha blending the Gaussians in front-to-back order:
\begin{equation}
\mathbf{C}=\sum_{i}w_i\mathbf{c}_i\,, \  \text{where}\quad  w_i = \alpha_i\mathcal{G}_i\prod_{j=1}^{i-1}(1-\alpha_j\mathcal{G}_j)\,,
\label{eq:alpha-blending}
\end{equation}
where $\mathbf{c}_i$ and $\alpha_i$ are the color and opacity of the $i$-th Gaussian, and $\mathcal{G}_i$ is the value of its projected 2D Gaussian kernel at the pixel center.

To better align the scene representation with surfaces, recent work has proposed 2D Gaussian Splatting (2DGS) \cite{2dgs}. Instead of 3D primitives, 2DGS models the scene as a set of 2D Gaussian surfels embedded in 3D space. This surface-aligned representation provides each primitive with a well-defined surface normal, typically derived from the orientation of the 2D disk. Furthermore, it mitigates the multi-view depth inconsistency issues that can arise from projecting 3D Gaussians, leading to a more geometrically accurate surface representation. Our work builds upon this 2DGS framework.

\subsection{Deferred Shading}

Deferred shading is a two-pass rendering technique that decouples geometry processing from lighting and material computations. In the first pass, known as the geometry pass, various attributes of the nearest surface, such as depth, normal, albedo, and roughness, are rendered into a set of intermediate 2D buffers, collectively called G-buffers. In the second pass, a shading program is executed for each pixel, using the information stored in the G-buffers to compute the final color. Recent works \cite{3dgs-dr,dihlmann2024subsurface,gi-gs,gs-id,wu2024deferredgs,ye2024progressive,gs-ror2,rgs-dr,yao2024reflective,geosplatting,spectre-gs,refgs,envgs} have successfully adapted this pipeline to Gaussian Splatting to efficiently render high-frequency, view-dependent effects. By performing complex shading calculations on a per-pixel basis rather than a per-Gaussian basis, deferred shading significantly enhances rendering quality and performance for complex materials.
\begin{figure*}[t]
    \centering
    \includegraphics[width=1.0\linewidth]{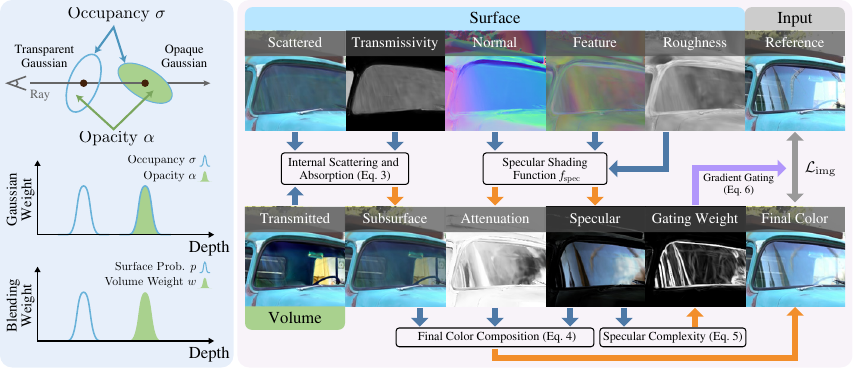}
    \caption{\textbf{Overview of RT-Splatting.} (Left) Transparent objects are represented by Gaussians with high geometric occupancy but low optical opacity, yielding strong contributions to surface aggregation while avoiding occlusion during volumetric compositing. (Right) Our hybrid rendering pipeline composites surface-based reflections from a deferred pass with volumetric transmission from a forward pass to produce the final color, and suppresses image-loss gradients flowing into the transmission branch according to the specular complexity of the corresponding pixels.}
    \label{fig:overview}
\end{figure*}

\section{Method}
\label{sec:method}

Our method is designed to reconstruct scenes with thin semi-transparent surfaces that exhibit both sharp reflections and clear transmission. We factorize the per-Gaussian opacity into geometric occupancy and optical opacity (\cref{sec:occupancy-opacity-factorization}), yielding a unified surface-volume representation that supports a hybrid pipeline for rendering reflections and transmission (\cref{sec:reflection-transmission-modeling}). To suppress floaters caused by residual reflection errors, we introduce Specular-Aware Gradient Gating (\cref{sec:specular-aware-gradient-gating}), and we finally describe the optimization details in \cref{sec:optimization}. An overview of the framework is shown in \cref{fig:overview}.

\subsection{Occupancy-Opacity Factorization}
\label{sec:occupancy-opacity-factorization}

The standard Gaussian Splatting pipeline uses a single opacity parameter for alpha blending, primarily to model optical occlusion. While recent deferred shading methods~\cite{3dgs-dr,refgs,envgs} have successfully repurposed this opacity to rasterize surface properties into G-buffers, this formulation conflates a Gaussian's geometric presence with its optical properties. This approximation is reasonable for opaque objects, but it fails fundamentally for semi-transparent surfaces, such as windows or plastic films. For these materials, the surface is geometrically solid (required for rendering sharp reflections) yet optically clear (allowing light transmission). A single opacity parameter cannot simultaneously satisfy these conflicting demands, leading to either blurry reflections or an opaque appearance.

To address this limitation, we factorize the standard per-Gaussian opacity into two physically motivated, learnable attributes. The \textit{geometric occupancy} $\sigma \in [0,1]$ encodes the probability that a ray interacts with the substance of the Gaussian. The \textit{optical opacity} $\alpha \in [0,1]$ then specifies the conditional probability that the ray is absorbed or scattered once such an interaction occurs. Their product $\alpha_{\text{eff}} = \sigma \alpha$ defines the effective opacity used for volumetric compositing in \cref{eq:alpha-blending}, meaning that optical attenuation only happens where the Gaussian is geometrically present. This factorization enables us to model transparent objects using Gaussians with high geometric occupancy but low optical opacity.

Our factorization naturally yields a probabilistic formulation for first-surface extraction, which is essential for deferred shading. Given a sequence of Gaussians along a ray sorted by depth, the expected value of any surface attribute $\mathbf{a}$ (\eg, normal or roughness) is computed as:
\begin{equation}
    \mathbf{A} = \sum_{i} p_i \mathbf{a}_i, \  \text{where}\quad  p_i = \sigma_i\mathcal{G}_i \prod_{j=1}^{i-1} (1 - \sigma_j\mathcal{G}_j)\,.
\label{eq:probabilistic-aggregation}
\end{equation}
Here, $p_i$ represents the probability that the $i$-th Gaussian is the first surface element with which the ray interacts. 

While mathematically analogous to standard alpha blending, our formulation provides a crucial reinterpretation: we treat the collection of Gaussians not as discrete, semi-transparent surfels, but as a unified, probabilistic representation of a single surface. This physically-grounded view justifies the application of deferred shading for high-frequency reflection modeling in Gaussian Splatting.

\subsection{Reflection-Transmission Modeling}
\label{sec:reflection-transmission-modeling}

To model the complex appearance of semi-transparent surfaces, which involves both high-frequency specular reflections and transmitted light, we propose a hybrid deferred-forward rendering framework.

Our framework begins with a deferred pass to handle high-frequency specular reflections on the first-hit surface. Leveraging our occupancy-opacity factorization, we first aggregate the expected surface properties into G-buffers using the probabilistic formulation in \cref{eq:probabilistic-aggregation}. Once the G-buffers are populated, a specular shading function $f_\text{spec}$ takes the view direction and a set of surface attributes, including normal $\mathbf{n}$, roughness $\rho$, and material feature $\mathbf{z}$, as input to compute the specular color $\mathbf{C}_\text{spec}$ for each pixel. This function is designed to reproduce complex, view-dependent specular effects, capturing reflections from the surrounding environment. For our implementation, we adopt a specular shading network architecture similar to that in Ref-GS~\cite{refgs}, which has proven effective for this task.

To capture the intrinsic appearance of materials like colored glass, which involves internal scattering and absorption, we introduce two additional surface attributes for each Gaussian: an intrinsic scattered color $\mathbf{C}_\text{scatter}$ and a transmissivity ratio $\tau\in[0,1]$. $\mathbf{C}_\text{scatter}$ represents light scattered back from within the material, while $\tau$ dictates the material's transmissivity by controlling the balance between this scattered light and the transmitted background light.

The background radiance itself, $\mathbf{C}_\text{trans}$, is computed with a concurrent forward pass. This pass operates like standard volumetric rendering, accumulating color from the opaque background scene. Crucially, it is accumulated with our effective opacity $\alpha_{\text{eff}}=\sigma\alpha$. This formulation allows the background scene to be correctly accumulated without being occluded by the transparent objects. We group all radiance that travels inside the material, including both transmitted and scattered components, into a subsurface-transport component in our formulation:
\begin{equation}
    \mathbf{C}_\text{sub} = \tau \mathbf{C}_\text{trans} + (1-\tau) \mathbf{C}_\text{scatter}\,.
    \label{eq:subsurface-color}
\end{equation}

Finally, we combine the specular reflection $\mathbf{C}_\text{spec}$ and the subsurface-transport component $\mathbf{C}_\text{sub}$ to produce the final pixel color. A purely physics-based blend using Fresnel equations is often broken in practice by tone-mapping and other nonlinear camera responses, and fails to capture our key perceptual observation: transmitted details are clearly visible through faint reflections but are suppressed or even masked by strong specular highlights. To model this dynamic effect, we augment our specular shading function to also output an attenuation factor $\beta \in [0, 1]$ that directly modulates the subsurface-transport component. The final color $\mathbf{C}$ is then computed as:
\begin{equation}
    \mathbf{C} = \mathbf{C}_\text{spec}+\beta\mathbf{C}_\text{sub}\,.
    \label{eq:final-color}
\end{equation}
Unlike previous methods~\cite{gao2024planar,nerfren,zhang2024refgaussian} that modulate the reflection component, our approach attenuates the transmitted component, which provides a direct and stable mechanism to model the suppression of background light.

\subsection{Specular-Aware Gradient Gating}
\label{sec:specular-aware-gradient-gating}

While our hybrid deferred-forward rendering pipeline cleanly separates how reflection and transmission are rendered, jointly optimizing both branches remains challenging. High-frequency specular reflections are inherently difficult to model perfectly, leaving residual discrepancies between the rendered reflections and the ground-truth observations. During backpropagation, gradients induced by these residuals can be erroneously routed into the transmission branch, which then compensates by hallucinating spurious floaters behind the surface and degrading the clarity of the transmitted background.

To mitigate this erroneous gradient flow, we introduce a specular-aware gradient gating mechanism. Our key insight is that this incorrect compensation primarily occurs in image regions with high-frequency specular details. We identify these regions by using the local variance of the specular component, $\mathbf{C}_\text{spec}$, to estimate its complexity. For each pixel $x$, we compute a gating weight $g(x)$ over a small neighboring patch $\mathcal{N}(x)$:
\begin{equation}
g(x) = \exp \left(-k \cdot \text{Var}_{p \in \mathcal{N}(x)} [\mathbf{C}_\text{spec}(p)] \right)\,,
\label{eq:specular-aware-gradient-gating}
\end{equation}
where $\text{Var}(\cdot)$ is the variance operator and $k$ is a hyperparameter controlling the gate's sensitivity.

During the backward pass, this gating weight modulates the gradients flowing to the transmission branch. Specifically, we apply $g(x)$ to scale the gradient of the image loss $\mathcal{L}_\text{img}$ that backpropagates through the transmitted background color $\mathbf{C}_\text{trans}$:
\begin{equation}
    \frac{\partial \mathcal{L}_\text{img}}{\partial \mathbf{C}_\text{trans}(x)} \leftarrow g(x) \cdot \frac{\partial \mathcal{L}_\text{img}}{\partial \mathbf{C}_\text{trans}(x)}\,.
    \label{eq:specular-aware-gradient-gating-modulation}
\end{equation}

In other words, this specular-aware gradient gating attenuates gradients at pixels dominated by complex specular patterns, but does not completely block supervision of the background scene behind the semi-transparent surface. At viewpoints and pixels where specular reflections are simple or weak, $g(x)$ remains close to one, so the transmitted background continues to receive full supervision through the transparent interface. This preserves a valid optimization path for the background geometry and appearance.

\begin{figure*}[t]
    \centering
    \includegraphics[width=1.0\linewidth]{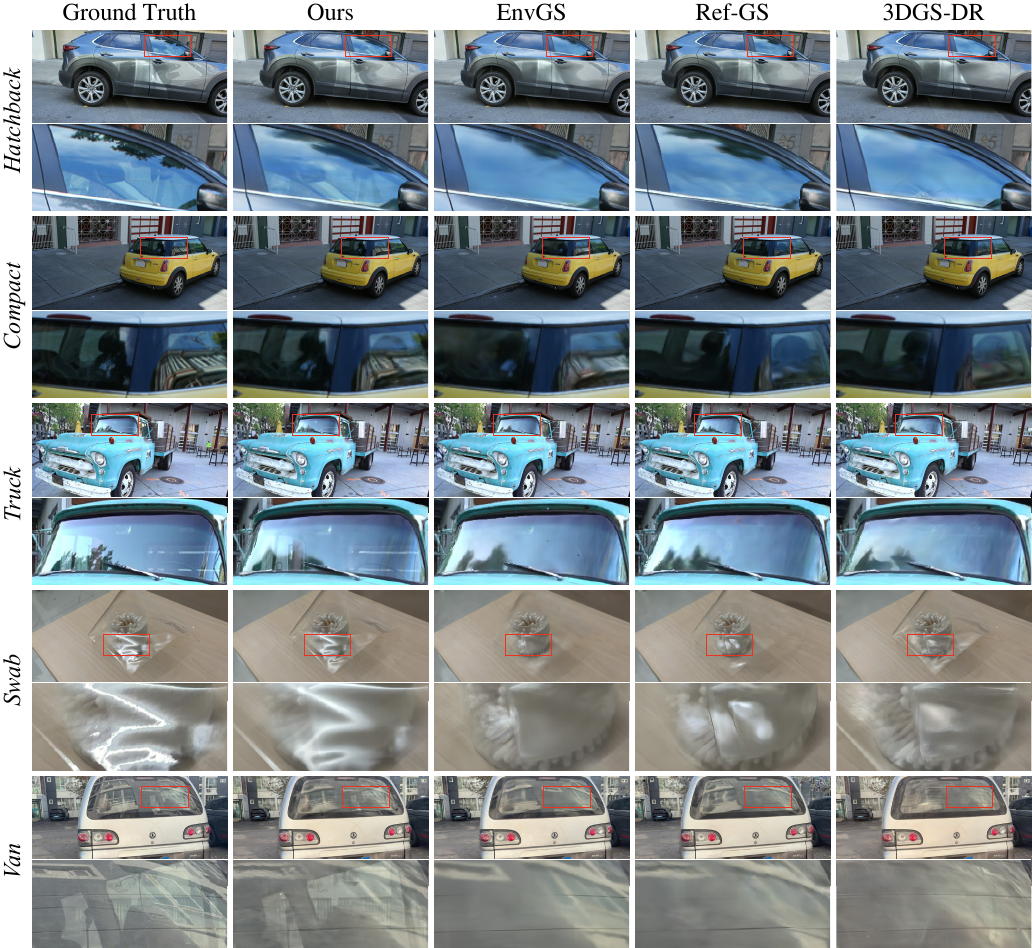}
    \caption{\textbf{Qualitative comparisons on test-set views of real-world scenes.} Our method significantly improves rendering quality over previous approaches, simultaneously yielding sharper reflections and clearer transmissions in semi-transparent regions.}
    \label{fig:compare}
    \vspace{-8pt}
\end{figure*}

\subsection{Optimization}
\label{sec:optimization}

\noindent\textbf{Transparent mask regularization.} Our occupancy-opacity factorization introduces a specific ambiguity: Gaussians with high geometric occupancy but near-zero optical opacity can exist anywhere in the scene without affecting the final rendered color. This is particularly problematic in diffuse regions lacking strong specular cues, where these unconstrained ``ghost'' geometries can accumulate, corrupting the surface representation and destabilizing the optimization process.

To resolve this ambiguity, we introduce a transparent mask loss that provides explicit supervision for the optical opacity of Gaussians. We leverage a transparent mask $\mathbf{M}$, obtained from the pre-trained SAM2 model~\cite{sam2,sam} to provide additional supervision. During the deferred pass, we aggregate the expected optical opacity $\alpha$ of the first-hit surface into G-buffers. We then supervise this opacity map with a binary cross-entropy (BCE) loss, encouraging it to match the inverted semantic mask:
\begin{equation}
    \mathcal{L}_{\text{mask}}=\text{BCE}(1-\mathbf{M}, \alpha)\,.
    \label{eq:transparent-mask-regularization}
\end{equation}

\noindent\textbf{Joint optimization.} We perform a joint optimization of all system components, simultaneously refining the Gaussian primitives, their factorized opacities, and the shading function. Unlike prior works~\cite{tsgs,transparentgs,bemana2022eikonal} that use the transparent mask to segment the scene for separate processing, our approach integrates the mask solely as a regularization. This joint optimization is crucial as it makes our method applicable to complex scenes where the background is exclusively visible through the transparent surfaces.
\section{Experiments}
\begin{table*}[t]
    \caption{\setlength{\fboxsep}{1pt}\textbf{Quantitative results on scenes from Ref-Real~\cite{refnerf}, NeRF-Casting~\cite{nerfcasting}, EnvGS~\cite{envgs} and T\&T~\cite{tandt}.} We report PSNR, SSIM~\cite{ssim}, and LPIPS~\cite{lpips} on entire images and over transparent regions, together with rendering speed (FPS) and training time. $\uparrow$ ($\downarrow$) indicates higher (lower) is better. We mark the \colorbox{tabfirst}{best}, the \colorbox{tabsecond}{second best}, and the \colorbox{tabthird}{third best} results in each column.}
    \label{tab:avg-baseline}
    \centering
    \footnotesize
    \begin{tabular}{l ccc|ccc|c|c}
        \toprule
        \multirow{2.5}{*}{\bfseries Methods} & \multicolumn{3}{c|}{\bfseries Entire Image} & \multicolumn{3}{c|}{\bfseries Transparent Region} & \multirow{2.5}{*}{\bfseries FPS $\uparrow$} & \multirow{2.5}{*}{\bfseries Training Time $\downarrow$} \\
        \cmidrule(lr){2-4} \cmidrule(lr){5-7}
        & PSNR $\uparrow$ & SSIM $\uparrow$ & LPIPS $\downarrow$ & PSNR $\uparrow$ & SSIM $\uparrow$ & LPIPS $\downarrow$ & \\
        \midrule
        3DGS~\cite{3dgs}&26.493&0.816&\cellsecond0.181&37.673&\cellsecond0.990&\cellsecond0.012&\textbf{218.95}&\textbf{0.3h}\\
        2DGS~\cite{2dgs}&26.384&0.817&0.197&37.333&\cellsecond0.990&\cellsecond0.012&208.82&\textbf{0.3h}\\
        GShader~\cite{gaussianshader}&25.778&0.806&0.203&36.797&0.988&0.014&24.59&1.3h\\
        3DGS-DR~\cite{3dgs-dr}&26.597&0.816&0.190&\cellthird37.890&\cellsecond0.990&\cellsecond0.012&119.62&0.8h\\
        Ref-GS~\cite{refgs}&\cellthird26.599&\cellthird0.819&0.188&37.761&\cellthird0.989&\cellthird0.013&38.41&0.8h\\
        EnvGS~\cite{envgs}&\cellsecond27.141&\cellsecond0.821&\cellthird0.182&\cellsecond37.953&\cellsecond0.990&\cellsecond0.012&18.31&2.9h\\
        Ours&\cellfirst27.490&\cellfirst0.831&\cellfirst0.167&\cellfirst39.765&\cellfirst0.992&\cellfirst0.010&33.28&0.9h\\
        \bottomrule
    \end{tabular}
    \vspace{-8pt}
\end{table*}

\subsection{Implementation Details}

We implement RT-Splatting in PyTorch~\cite{pytorch} building upon the 2DGS framework~\cite{2dgs}. 
The hyperparameters for the shading function in our deferred pass are kept consistent with those in Ref-GS~\cite{refgs}. Please refer to the supplementary materials for more details.

\subsection{Datasets and Metrics}

We evaluate our method on several real-world datasets that prominently feature the coexistence of high-frequency specular reflections and clear transmission on semi-transparent surfaces. From public benchmarks, we select six scenes: \textit{Sedan} and \textit{Toycar} from Ref-Real~\cite{refnerf}, \textit{Compact} and \textit{Hatchback} from NeRF-Casting~\cite{nerfcasting}, \textit{Audi} from EnvGS~\cite{envgs}, and \textit{Truck} from T\&T~\cite{tandt}. We additionally captured two real-world scenes, \textit{Van} and \textit{Swab}, using a smartphone camera to capture $220\sim 240$ views for each scene.

We report PSNR, SSIM~\cite{ssim}, and LPIPS~\cite{lpips} measured on both entire images and transparent regions. Additional results are provided in the supplementary materials.

\subsection{Baseline Comparisons}

We conduct a comprehensive comparison of our method against several state-of-the-art Gaussian Splatting variants. The baselines include the foundational 3DGS~\cite{3dgs}, 2DGS~\cite{2dgs}, and methods specifically designed for reflective surfaces. Among these are GaussianShader~\cite{gaussianshader}, as well as the recent deferred shading-based approaches 3DGS-DR~\cite{3dgs-dr}, Ref-GS~\cite{refgs}, and EnvGS~\cite{envgs}. All baseline models are trained using their publicly available codebases and configurations.

\begin{table}[t]
    \caption{\textbf{Quantitative results on our self-captured scenes.}}
    \label{tab:avg-self}
    \setlength{\tabcolsep}{4pt}
    \centering
    \footnotesize
    \resizebox{1.0\linewidth}{!}{
    \begin{tabular}{l ccc | ccc}
        \toprule
        \multirow{2.5}{*}{\bfseries Methods} & \multicolumn{3}{c|}{\bfseries Entire image} & \multicolumn{3}{c}{\bfseries Transparent region} \\
        \cmidrule(lr){2-4} \cmidrule(lr){5-7}
        & PSNR $\uparrow$ & SSIM $\uparrow$ & LPIPS $\downarrow$ & PSNR $\uparrow$ & SSIM $\uparrow$ & LPIPS $\downarrow$ \\
        \midrule
        3DGS~\cite{3dgs} & \cellsecond27.507 & \cellsecond0.863 & \cellsecond0.213 & \cellsecond32.567 & \cellthird0.964 & \cellsecond0.048 \\
        2DGS~\cite{2dgs} & 26.675 & 0.857 & 0.243 & 31.923 & \cellsecond0.965 & 0.053 \\
        GShader~\cite{gaussianshader} & 21.133 & 0.801 & 0.367 & 26.995 & 0.944 & 0.084 \\
        3DGS-DR~\cite{3dgs-dr} & 26.134 & \cellthird0.858 & \cellthird0.229 & \cellthird32.014 & \cellthird0.964 & \cellthird0.049 \\
        Ref-GS~\cite{refgs} & 26.301 & 0.851 & 0.240 & 31.646 & 0.962 & 0.056 \\
        EnvGS~\cite{envgs} & \cellthird26.847 & 0.847 & 0.260 & 31.726 & 0.963 & 0.057 \\
        Ours & \cellfirst28.780 & \cellfirst0.871 & \cellfirst0.197 & \cellfirst35.490 & \cellfirst0.970 & \cellfirst0.042 \\
        \bottomrule
    \end{tabular}
    }
    \vspace{-6pt}
\end{table}

We present quantitative results on both public benchmarks and our self-captured scenes in \cref{tab:avg-baseline} and \cref{tab:avg-self}. The results consistently show that our method outperforms all baselines across all evaluated metrics. This performance advantage is particularly significant when evaluating the transparent regions, highlighting our model's enhanced capability in these challenging areas. Notably, our method achieves real-time rendering speeds and maintains a competitive training time, proving its efficiency and practical applicability.

Qualitative comparisons presented in \cref{fig:compare} further demonstrate our approach's unique capability to faithfully render both sharp reflection details and clear transmitted light simultaneously. Existing methods struggle with the inherent ambiguity between reflection and transmission on semi-transparent surfaces. As illustrated, they often fail to reconstruct sharp reflection details, as the optimization is compromised by the underlying transmitted light. Conversely, attempts to model strong reflections typically result in the surface being rendered as opaque, thereby sacrificing transmission clarity and occluding the background scene entirely.

\subsection{Ablation Studies}

\begin{table}[t]
    \caption{
        \textbf{Ablation study of our model in transparent regions.} We mark the \textbf{best} results in each column.
    }
    \label{tab:avg-ablation}
    \centering 
    \footnotesize  
  
    \begin{tabular}{p{2.5cm}ccc}
        \toprule
        & PSNR $\uparrow$ & SSIM $\uparrow$ & LPIPS $\downarrow$ \\
        \midrule
        \textit{w/o} occupancy &36.919&0.9885&0.0113 \\
        \textit{w/o} joint optimization &36.288&0.9876&0.0120 \\
        \textit{w/o} scattering &37.597&0.9897&0.0102 \\
        \textit{w/o} attenuation &37.541&0.9897&0.0102 \\
        \textit{w/o} gating &37.754&0.9899&0.0101 \\
        \textit{w/o} $\mathcal{L}_{\text{mask}}$ &37.167&0.9894&0.0106 \\
        \midrule
        Ours &\textbf{37.983}&\textbf{0.9901}&\textbf{0.0095} \\
        \bottomrule
    \end{tabular}

    \vspace{-4pt}
\end{table}

\begin{figure*}[t]
    \centering
    \includegraphics[width=1.0\linewidth]{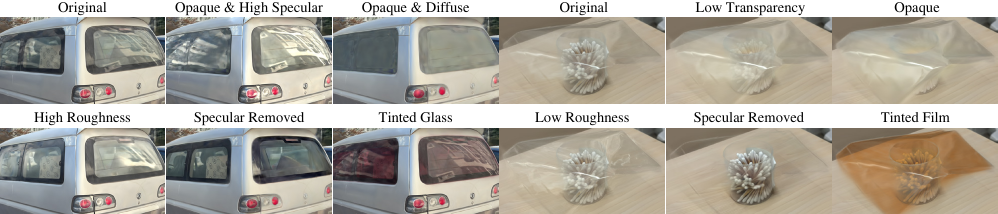}
    \caption{\textbf{Scene Editing.} Left: edited appearances of car windows. Right: edited appearances of a plastic film.}
    \label{fig:editing}
    \vspace{-8pt}
\end{figure*}

\begin{figure}[t]
    \centering
    \includegraphics[width=1.0\linewidth]{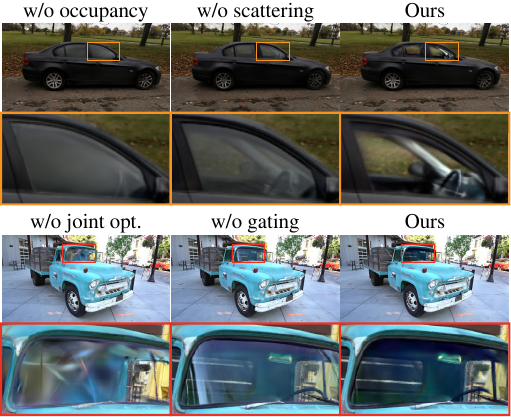}
    \caption{\textbf{Decomposed transmission components across ablation settings.}}
    \label{fig:ablation}
    \vspace{-8pt}
\end{figure}

We validate the effectiveness of our key components through ablation studies on \textit{Sedan} and \textit{Truck}. Quantitative and qualitative comparisons are presented in \cref{tab:avg-ablation} and \cref{fig:ablation}, respectively.

\noindent\textbf{Occupancy-opacity factorization.} The ``w/o occupancy'' variant removes our occupancy-opacity factorization, forcing a single opacity parameter to model both geometric occupancy and optical opacity. As shown in \cref{tab:avg-ablation} and \cref{fig:ablation}, this creates a conflict where achieving sharp reflections requires high opacity, which in turn severely compromises transmission clarity, leading to a more occluded background.

\noindent\textbf{Joint optimization.} The ``w/o joint optimization'' variant separates the training of the reflection and transmission components. As shown in \cref{tab:avg-ablation} and \cref{fig:ablation}, this completely prevents the reconstruction of the interior of the truck, which is exclusively visible through its windows.

\noindent\textbf{Internal scattering and absorption.} The ``w/o scattering'' variant removes the components responsible for the material's intrinsic appearance, namely the scattered color $\mathbf{C}_\text{scatter}$ and the transmissivity ratio $\tau$. This forces the model to bake the material's intrinsic appearance into the volumetric background scene, resulting in a much darker transmission, as shown in \cref{fig:ablation} and \cref{tab:avg-ablation}.

\noindent\textbf{Subsurface attenuation.} The ``w/o attenuation'' variant removes the learnable attenuation factor $\beta$. This fails to model the view-dependent suppression of subsurface component, resulting in reduced rendering quality as demonstrated in \cref{tab:avg-ablation}.

\noindent\textbf{Specular-aware gradient gating.} The ``w/o gating'' variant disables our Specular-Aware Gradient Gating mechanism discussed in \cref{sec:specular-aware-gradient-gating}. This causes the transmission branch to generate visual artifacts near the transparent surface as shown in \cref{fig:ablation} and \cref{tab:avg-ablation}.

\noindent\textbf{Transparent mask regularization.} The ``w/o $\mathcal{L}_{\text{mask}}$'' variant removes the transparent mask regularization. This leads to unstable optimization and degraded surface quality as demonstrated in \cref{tab:avg-ablation}.

\subsection{Applications}

Our method's reflection-transmission decomposition naturally facilitates a variety of powerful scene editing applications. As illustrated in \cref{fig:editing}, we can independently manipulate surface attributes by adjusting roughness, changing its transparency level, removing specular reflections, or even altering the material's tint. This demonstrates the effectiveness and intuitive control offered by our decoupled representation.

\section{Conclusion}

We presented RT-Splatting, a framework for jointly modeling high-fidelity reflections and clear transmissions on semi-transparent surfaces. By disentangling each Gaussian primitive's geometric occupancy from its optical opacity, a single set of Gaussians supports a hybrid renderer that simultaneously interprets the scene as a reflective surface for sharp specular highlights and as a transmissive volume for clear background content. Furthermore, we introduce a specular-aware gradient gating mechanism to mitigate optimization ambiguities between the reflection and transmission components. Together, these designs enable RT-Splatting to achieve state-of-the-art performance on challenging scenes where reflection and transmission are strongly coupled.

A limitation of RT-Splatting is that it is designed for thin semi-transparent surfaces, as it does not explicitly model refraction or multiple light bounces. Future work could explore extending our framework to handle thicker refractive media and multi-bounce light transport, such as in water or solid glass objects.
\clearpage

\section*{Acknowledgments}
This work was supported by the Beijing Natural Science Foundation under Grant No. L247029, and National Natural Science Foundation of China (NSFC) under Grant No. 62371009.

{
    \small
    \bibliographystyle{ieeenat_fullname}
    \bibliography{main}
}

\clearpage
\renewcommand{\thesection}{\Alph{section}}
\setcounter{page}{1}
\setcounter{section}{0}
\maketitlesupplementary

In the supplementary material, we provide additional implementation details of our method (\cref{sec:appendix:implementation-details}). We also present an ablation study that examines the sensitivity of Specular-Aware Gradient Gating to the gating strength $k$ (\cref{sec:ablation-study-gating-k}). Finally, we report a per-scene breakdown of quantitative metrics and include further qualitative results (\cref{sec:appendix:additional-results}).

\section{Implementation Details}
\label{sec:appendix:implementation-details}

We strictly follow the evaluation protocols established in 3DGS-DR~\cite{3dgs-dr} and Ref-GS~\cite{refgs}. Consistent with these methods, we adopt the spherical domain strategy to define the foreground region of interest. Furthermore, we adopt their data preprocessing standards by applying consistent image downsampling factors across all experiments to ensure a fair comparison. All experiments are conducted on a single NVIDIA RTX 4090 GPU.

\subsection{Density Control Strategy}
Our density control strategy adapts the standard 2DGS~\cite{2dgs} scheme to align with our proposed occupancy-opacity factorization. While standard 2DGS relies on opacity resets to regulate the number of primitives and prevent floaters, our decoupled representation requires managing the Gaussian primitives based on both geometric occupancy and optical opacity. We therefore introduce an interleaved reset schedule: instead of the standard opacity reset every 3,000 iterations, we perform resets every 1,500 iterations, alternating between geometric occupancy $\sigma$ and optical opacity $\alpha$. Furthermore, we perform pruning based on geometric occupancy $\sigma$ rather than optical opacity $\alpha$ to prevent the erroneous removal of semi-transparent structures.

\subsection{Specular-Aware Gradient Gating}
For the local variance calculation in \cref{eq:specular-aware-gradient-gating}, we use a $3 \times 3$ window size. The hyperparameter $k$ is set to $4$ based on our sensitivity analysis in \cref{sec:ablation-study-gating-k}. In practice, the gradient modulation described in \cref{eq:specular-aware-gradient-gating-modulation} is implemented via a partial stop-gradient mechanism. Let $\texttt{sg}(\cdot)$ denote the stop-gradient operator. The gated transmission color $\tilde{\mathbf{C}}_{\text{trans}}$ used for final composition is defined as:
\begin{equation}
    \tilde{\mathbf{C}}_{\text{trans}} = (1 - g)\cdot\texttt{sg}(\mathbf{C}_{\text{trans}}) + g\cdot\mathbf{C}_{\text{trans}}\,.
\end{equation}
This formulation ensures $\tilde{\mathbf{C}}_{\text{trans}} \equiv \mathbf{C}_{\text{trans}}$ during the forward pass, while scaling the backward gradients by $g$.

\subsection{Losses}

Following 2DGS~\cite{2dgs}, we minimize the normal consistency loss $\mathcal{L}_{\text{n}}$ to enforce geometric alignment between rendered normals and depth gradients. For appearance supervision $\mathcal{L}_{\text{img}}$, we augment the standard $\mathcal{L}_1$ and D-SSIM~\cite{ssim} losses with the perceptual loss $\mathcal{L}_{\text{perc}}$ from EnvGS~\cite{envgs}, with $\lambda=0.2$ and $\lambda_{\text{perc}}=0.01$:
\begin{equation}
\mathcal{L}_{\text{img}} = (1 - \lambda)\mathcal{L}_{1} + \lambda\mathcal{L}_{\text{D-SSIM}} + \lambda_{\text{perc}}\mathcal{L}_{\text{perc}}\,.
\end{equation}

Combining these with our transparent mask regularization $\mathcal{L}_{\text{mask}}$ in \cref{eq:transparent-mask-regularization}, the total loss is given as:
\begin{equation}
    \mathcal{L} = \mathcal{L}_{\text{img}} + \lambda_{\text{n}}\mathcal{L}_{\text{n}} + \lambda_{\text{mask}}\mathcal{L}_{\text{mask}}\,.
\end{equation}
We empirically set $\lambda_{\text{n}} = 0.05$ and $\lambda_{\text{mask}} = 0.01$.

\begin{table}[t]
    \caption{Sensitivity analysis of the gating strength hyperparameter $k$, averaged over transparent regions of all scenes.}
    \label{tab:gating-k}
    \centering
    \footnotesize
    \begin{tabular}{cccc}
        \toprule
        $k$ & PSNR$\uparrow$ & SSIM$\uparrow$ & LPIPS$\downarrow$ \\
        \midrule
        0   & 38.574 & 0.9861 & 0.0179 \\
        1   & 38.436 & 0.9860 & 0.0185 \\
        2   & 38.631 & 0.9862 & 0.0178 \\
        4   & \textbf{38.696} & \textbf{0.9865} & \textbf{0.0175} \\
        8   & 38.608 & 0.9863 & 0.0176 \\
        16  & 38.523 & 0.9864 & 0.0179 \\
        32  & 38.417 & 0.9863 & 0.0180 \\
        \bottomrule
    \end{tabular}
\end{table}

\section{Additional Ablation Study}
\label{sec:ablation-study-gating-k}

To evaluate the impact of our Specular-Aware Gradient Gating mechanism, we conduct a sensitivity analysis on the gating strength hyperparameter $k$ defined in \cref{eq:specular-aware-gradient-gating}. The quantitative results, which are averaged over the transparent regions across all eight scenes, are summarized in \cref{tab:gating-k}. The performance peaks at $k=4$, which offers the optimal trade-off between suppressing specular artifacts and preserving transmission details.

\section{Additional Results}
\label{sec:appendix:additional-results}

\begin{figure*}[t]
    \centering
    \includegraphics[width=1.0\linewidth]{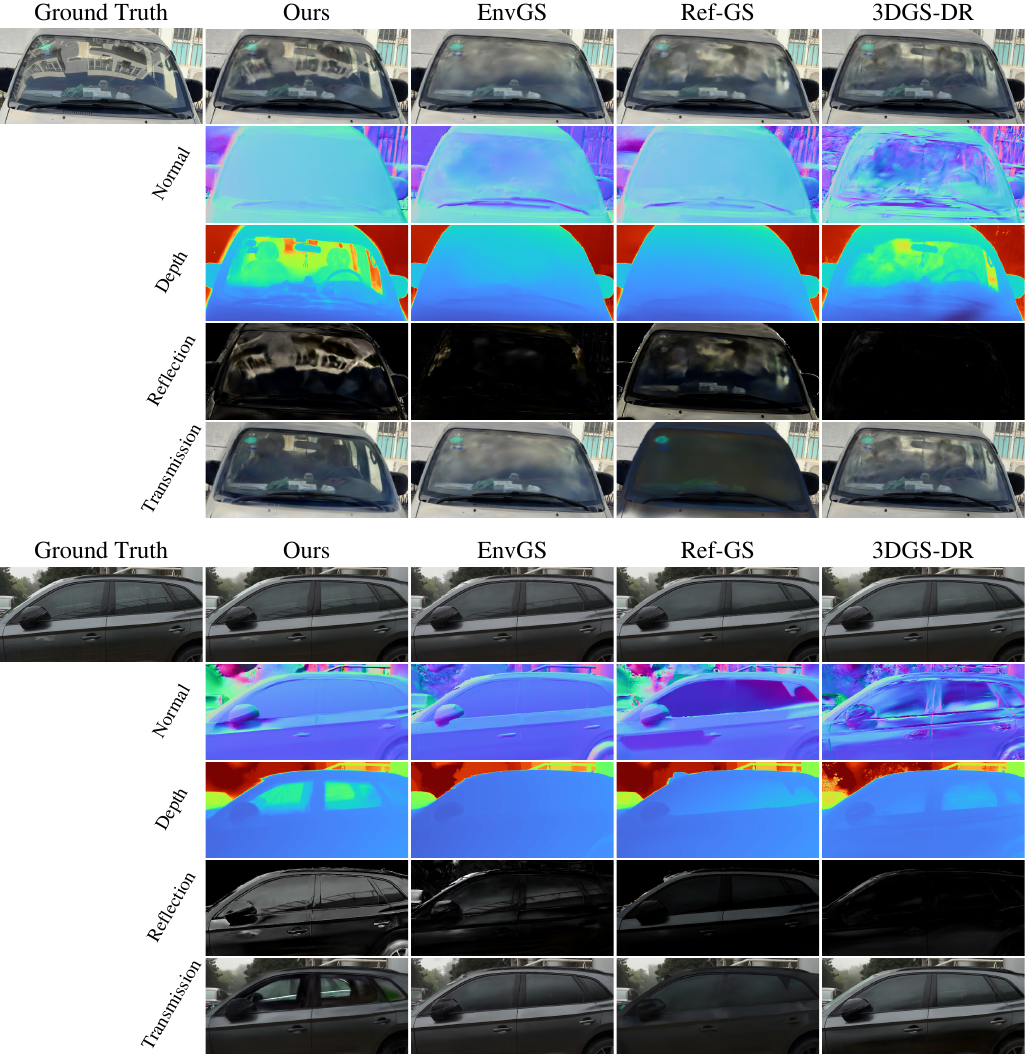}
    \caption{\textbf{Visual decomposition on real-world scenes.} We visualize the decomposed components of our method compared to baselines. For our method, \textit{Normal} captures the surface geometry, while \textit{Depth} corresponds to the volumetric accumulation. Since baseline methods do not explicitly model semi-transparent transmission, we visualize their diffuse component in the \textit{Transmission} row. Our method achieves high-fidelity separation of the \textit{Reflection} and \textit{Transmission} layers.}
    \label{fig:appendix_decomposition}
\end{figure*}

Tables~\ref{tab:appendix_per_scene_entire} and~\ref{tab:appendix_per_scene_transparent} report per-scene quantitative metrics for entire images and transparent regions, respectively. In addition, \cref{fig:appendix_decomposition} presents further qualitative results to demonstrate the decomposition capabilities of our method. In contrast, baseline methods often fail to simultaneously reconstruct surface reflections and the transmission behind the surface, recovering only one component at the expense of the other. We also recommend viewing the supplementary video to best appreciate the temporal consistency and visual quality of our results.

\begin{table*}[t!]
    \centering
    \small
    \begin{tabular}{lcccccccc}
        \toprule
        \multirow{2}{*}{\bfseries Methods} & \multicolumn{8}{c}{\bfseries PSNR $\uparrow$} \\
                                           & \textit{Sedan} & \textit{Toycar} & \textit{Compact} & \textit{Hatchback} & \textit{Audi} & \textit{Truck} & \textit{Van} & \textit{Swab} \\

        \midrule
        3DGS~\cite{3dgs} & 26.038 & 23.779 & 29.506 & 26.556 & 27.683 & \cellthird25.394 & \cellsecond23.436 & \cellsecond31.578 \\
        2DGS~\cite{2dgs} & 25.467 & 24.012 & 29.602 & 26.623 & 27.487 & 25.113 & 23.317 & 30.032 \\
        GShader~\cite{gaussianshader} & 25.709 & 23.888 & 28.542 & 25.880 & 26.913 & 23.738 & 22.189 & 20.077 \\
        3DGS-DR~\cite{3dgs-dr} & 26.117 & 24.123 & 29.232 & \cellthird27.088 & 27.778 & 25.246 & \cellthird23.385 & 28.882 \\
        Ref-GS~\cite{refgs} & \cellthird26.397 & \cellthird24.176 & \cellsecond29.706 & 26.556 & \cellthird27.858 & 24.902 & 22.351 & 30.251 \\
        EnvGS~\cite{envgs} & \cellsecond26.790 & \cellsecond24.637 & \cellthird29.703 & \cellsecond27.457 & \cellsecond28.686 & \cellsecond25.571 & 23.043 & \cellthird30.651 \\
        \midrule
        Ours & \cellfirst27.071 & \cellfirst24.794 & \cellfirst30.598 & \cellfirst27.619 & \cellfirst29.011 & \cellfirst25.848 & \cellfirst23.744 & \cellfirst33.817 \\
        \bottomrule
            \addlinespace[8pt]
        \toprule
        \multirow{2}{*}{\bfseries Methods} & \multicolumn{8}{c}{\bfseries SSIM $\uparrow$} \\
                                           & \textit{Sedan} & \textit{Toycar} & \textit{Compact} & \textit{Hatchback} & \textit{Audi} & \textit{Truck} & \textit{Van} & \textit{Swab} \\

        \midrule
        3DGS~\cite{3dgs} & 0.768 & 0.637 & \cellsecond0.895 & 0.845 & 0.874 & \cellsecond0.878 & \cellsecond0.778 & \cellsecond0.949 \\
        2DGS~\cite{2dgs} & 0.770 & 0.653 & 0.888 & \cellsecond0.848 & \cellthird0.875 & 0.870 & 0.768 & \cellthird0.945 \\
        GShader~\cite{gaussianshader} & 0.760 & 0.649 & 0.885 & 0.840 & 0.860 & 0.844 & 0.747 & 0.855 \\
        3DGS-DR~\cite{3dgs-dr} & 0.768 & 0.655 & 0.883 & \cellthird0.846 & 0.869 & 0.873 & \cellthird0.773 & 0.943 \\
        Ref-GS~\cite{refgs} & \cellsecond0.778 & \cellthird0.658 & \cellfirst0.898 & 0.845 & 0.864 & 0.871 & 0.758 & 0.943 \\
        EnvGS~\cite{envgs} & \cellthird0.777 & \cellsecond0.667 & 0.878 & \cellsecond0.848 & \cellsecond0.881 & \cellthird0.876 & 0.753 & 0.942 \\
        \midrule
        Ours & \cellfirst0.790 & \cellfirst0.684 & \cellthird0.893 & \cellfirst0.851 & \cellfirst0.887 & \cellfirst0.881 & \cellfirst0.785 & \cellfirst0.958 \\
        \bottomrule
            \addlinespace[8pt]
        \toprule
        \multirow{2}{*}{\bfseries Methods} & \multicolumn{8}{c}{\bfseries LPIPS $\downarrow$} \\
                                           & \textit{Sedan} & \textit{Toycar} & \textit{Compact} & \textit{Hatchback} & \textit{Audi} & \textit{Truck} & \textit{Van} & \textit{Swab} \\

        \midrule
        3DGS~\cite{3dgs} & \cellsecond0.205 & 0.237 & \cellsecond0.147 & 0.196 & \cellthird0.150 & \cellsecond0.148 & \cellsecond0.233 & \cellsecond0.193 \\
        2DGS~\cite{2dgs} & 0.234 & 0.238 & 0.175 & 0.198 & 0.164 & 0.174 & 0.266 & 0.219 \\
        GShader~\cite{gaussianshader} & 0.226 & 0.258 & 0.167 & \cellthird0.195 & 0.177 & 0.192 & 0.282 & 0.453 \\
        3DGS-DR~\cite{3dgs-dr} & 0.209 & 0.240 & 0.163 & 0.204 & 0.160 & 0.166 & \cellthird0.255 & \cellthird0.203 \\
        Ref-GS~\cite{refgs} & \cellsecond0.205 & \cellthird0.236 & \cellthird0.149 & 0.196 & 0.183 & \cellthird0.156 & 0.263 & 0.217 \\
        EnvGS~\cite{envgs} & \cellthird0.207 & \cellsecond0.233 & 0.167 & \cellsecond0.183 & \cellsecond0.139 & 0.163 & 0.267 & 0.253 \\
        \midrule
        Ours & \cellfirst0.188 & \cellfirst0.227 & \cellfirst0.141 & \cellfirst0.177 & \cellfirst0.137 & \cellfirst0.129 & \cellfirst0.215 & \cellfirst0.179 \\
        \bottomrule
    \end{tabular}

    \caption{\textbf{Per-scene metrics on entire images.} We report PSNR, SSIM, and LPIPS on scenes  from Ref-Real~\cite{refnerf}, NeRF-Casting~\cite{nerfcasting}, EnvGS~\cite{envgs}, T\&T~\cite{tandt}, and our self-captured scenes.}
    \label{tab:appendix_per_scene_entire}
\end{table*}

\begin{table*}[t!]
    \centering
    \small
    \begin{tabular}{lcccccccc}
        \toprule
        \multirow{2}{*}{\bfseries Methods} & \multicolumn{8}{c}{\bfseries PSNR $\uparrow$} \\
                                           & \textit{Sedan} & \textit{Toycar} & \textit{Compact} & \textit{Hatchback} & \textit{Audi} & \textit{Truck} & \textit{Van} & \textit{Swab} \\

        \midrule
        3DGS~\cite{3dgs} & 35.514 & 40.462 & 37.212 & \cellthird37.543 & 38.742 & \cellthird36.567 & \cellthird30.652 & \cellsecond34.481 \\
        2DGS~\cite{2dgs} & 35.315 & 40.379 & \cellthird37.276 & 36.555 & 37.932 & 36.541 & \cellsecond30.720 & 33.126 \\
        GShader~\cite{gaussianshader} & 35.394 & \cellthird40.545 & 35.194 & 36.017 & 38.300 & 35.334 & 28.836 & 25.155 \\
        3DGS-DR~\cite{3dgs-dr} & 36.489 & 40.137 & \cellsecond37.301 & 37.241 & \cellsecond39.518 & \cellsecond36.655 & 30.547 & 33.480 \\
        Ref-GS~\cite{refgs} & \cellsecond37.037 & 40.338 & 37.212 & \cellthird37.543 & 38.540 & 35.894 & 29.711 & \cellthird33.581 \\
        EnvGS~\cite{envgs} & \cellthird36.729 & \cellsecond40.638 & 36.777 & \cellsecond37.609 & \cellthird39.475 & 36.492 & 29.917 & 33.535 \\
        \midrule
        Ours & \cellfirst37.727 & \cellfirst40.899 & \cellfirst40.785 & \cellfirst38.927 & \cellfirst42.015 & \cellfirst38.238 & \cellfirst32.429 & \cellfirst38.551 \\
        \bottomrule
            \addlinespace[8pt]
        \toprule
        \multirow{2}{*}{\bfseries Methods} & \multicolumn{8}{c}{\bfseries SSIM $\uparrow$} \\
                                           & \textit{Sedan} & \textit{Toycar} & \textit{Compact} & \textit{Hatchback} & \textit{Audi} & \textit{Truck} & \textit{Van} & \textit{Swab} \\

        \midrule
        3DGS~\cite{3dgs} & 0.984 & \cellfirst0.995 & \cellsecond0.988 & \cellsecond0.991 & \cellsecond0.990 & \cellsecond0.991 & \cellthird0.949 & \cellsecond0.979 \\
        2DGS~\cite{2dgs} & 0.985 & \cellfirst0.995 & \cellsecond0.988 & \cellsecond0.991 & \cellthird0.989 & \cellsecond0.991 & \cellsecond0.951 & \cellthird0.978 \\
        GShader~\cite{gaussianshader} & 0.984 & \cellfirst0.995 & 0.984 & \cellthird0.990 & 0.988 & 0.988 & 0.945 & 0.944 \\
        3DGS-DR~\cite{3dgs-dr} & \cellthird0.986 & \cellfirst0.995 & \cellsecond0.988 & \cellsecond0.991 & \cellthird0.989 & \cellthird0.990 & \cellthird0.949 & \cellthird0.978 \\
        Ref-GS~\cite{refgs} & \cellsecond0.987 & \cellsecond0.994 & \cellsecond0.988 & \cellsecond0.991 & 0.988 & 0.989 & 0.947 & 0.976 \\
        EnvGS~\cite{envgs} & \cellthird0.986 & \cellfirst0.995 & \cellthird0.987 & \cellsecond0.991 & \cellsecond0.990 & 0.989 & \cellthird0.949 & 0.977 \\
        \midrule
        Ours & \cellfirst0.988 & \cellfirst0.995 & \cellfirst0.992 & \cellfirst0.993 & \cellfirst0.992 & \cellfirst0.992 & \cellfirst0.954 & \cellfirst0.986 \\
        \bottomrule
            \addlinespace[8pt]
        \toprule
        \multirow{2}{*}{\bfseries Methods} & \multicolumn{8}{c}{\bfseries LPIPS $\downarrow$} \\
                                           & \textit{Sedan} & \textit{Toycar} & \textit{Compact} & \textit{Hatchback} & \textit{Audi} & \textit{Truck} & \textit{Van} & \textit{Swab} \\

        \midrule
        3DGS~\cite{3dgs} & 0.016 & \cellfirst0.005 & \cellthird0.015 & \cellthird0.010 & \cellsecond0.018 & \cellsecond0.008 & \cellsecond0.053 & \cellsecond0.042 \\
        2DGS~\cite{2dgs} & 0.017 & \cellfirst0.005 & \cellsecond0.014 & \cellthird0.010 & 0.020 & \cellsecond0.008 & 0.056 & 0.049 \\
        GShader~\cite{gaussianshader} & 0.017 & \cellsecond0.006 & 0.019 & 0.011 & 0.020 & 0.010 & 0.061 & 0.107 \\
        3DGS-DR~\cite{3dgs-dr} & \cellthird0.015 & \cellfirst0.005 & \cellthird0.015 & \cellthird0.010 & \cellthird0.019 & \cellthird0.009 & \cellthird0.054 & \cellthird0.044 \\
        Ref-GS~\cite{refgs} & \cellsecond0.014 & \cellsecond0.006 & \cellthird0.015 & \cellthird0.010 & 0.022 & 0.010 & 0.060 & 0.052 \\
        EnvGS~\cite{envgs} & \cellthird0.015 & \cellsecond0.006 & 0.016 & \cellsecond0.009 & \cellsecond0.018 & 0.010 & 0.058 & 0.055 \\
        \midrule
        Ours & \cellfirst0.012 & \cellfirst0.005 & \cellfirst0.010 & \cellfirst0.008 & \cellfirst0.015 & \cellfirst0.007 & \cellfirst0.049 & \cellfirst0.034 \\
        \bottomrule
    \end{tabular}

    \caption{\textbf{Per-scene metrics on transparent regions.}}
    \label{tab:appendix_per_scene_transparent}
\end{table*}

\end{document}